\documentclass{article}
\usepackage{latexsym}
\usepackage{amsmath,amsthm,amsopn,amssymb}
\usepackage{bbding}
\usepackage{graphicx}
\usepackage{hyperref}
\usepackage{mathptmx}
\allowdisplaybreaks

\numberwithin{equation}{section}
\newtheorem{theorem}{Theorem}[section]
\newtheorem{lemma}{Lemma}[section]

\newtheorem{algorithm}{Algorithm}[section]

\makeatother

\begin{document}

\title{A Frobenius norm regularization method for  convolutional kernels to avoid unstable gradient problem
}

\author{
Pei-Chang Guo \thanks{ e-mail:peichang@cugb.edu.cn} \\
School of Science,
China University of Geosciences, Beijing, 100083, China\\
}

\date{}

\maketitle
\begin{abstract}
Convolutional neural network is a very important model of deep learning. It can help avoid the exploding/vanishing gradient problem and improve the generalizability of a neural network if the singular values of the Jacobian of a layer are bounded around $1$ in the training process.  We propose a new  penalty function for a convolutional kernel to let the singular values of the  corresponding transformation matrix are  bounded around $1$.  We show how to carry out the gradient type methods. The penalty is about the structured transformation matrix corresponding to a convolutional kernel. This provides a new regularization method about  the weights of convolutional layers.

\vspace{2mm} \noindent \textbf{Keywords}:  regularization, transformation matrix, convolutional kernel, generalizability, unstable gradient.
\end{abstract}

\section{Introduction}

 Convolution without the flip is an important arithmetic in the field of deep learning  \cite{dumoulin2018}. Depending on different strides and padding patterns, there are many different forms of convolution arithmetic\cite{dumoulin2018}. Without losing generality, in this paper we will consider the same convolution with unit strides.  Our objective is to penalize the kernel $K$ to let the singular values of the corresponding  transformation matrix $M$  be bounded around $1$.

 First we introduce the convolution arithmetic in deep learning,  which is different from the convolution in signal processing. When we refer to convolution in deep learning, only element-wise multiplication and addition are performed.  There is no reverse for the convolutional kernel in deep learning. We use $*$ to denote the convolution arithmetic in deep learning and $\ulcorner \cdot\urcorner$ is to round a number to the nearest integer greater than or equal to that number. If a convolutional kernel is a matrix $K\in \mathbb{R}^{k\times k}$  and the input is a matrix  $X\in \mathbb{R}^{N\times N}$,  each entry of the output $Y\in \mathbb{R}^{N \times N}$  is produced by
\begin{equation*}
    Y_{r,s}= (K*X)_{r,s}= \sum_{p\in \{1,\cdots,k\} }\sum_{q\in \{1,\cdots,k\}} X_{r-m+p,s-m+q}K_{p,q},
\end{equation*}
where $m=\ulcorner k/2\urcorner$, , and $X_{i,j}=0$ if $i\leq 0$ or $i> N$, or $j\leq 0$ or $j> N$.

In convolutional neural networks, usually there are multi-channels and a convolutional kernel is  represented by a 4 dimensional tensor.
If a convolutional kernel is a 4 dimensional tensor $K\in \mathbb{R}^{k\times k\times g  \times h}$  and the input is 3 dimensional tensor $X\in \mathbb{R}^{N\times N\times g}$, each entry of  the output $Y\in \mathbb{R}^{N \times N \times h}$  is produced by
\begin{equation*}
    Y_{r,s,c}= (K*X)_{r,s,c}=\sum_{d\in \{1,\cdots,g\}} \sum_{p\in \{1,\cdots,k\} }\sum_{q\in \{1,\cdots,k\}} X_{r-m+p,s-m+q,d}K_{p,q,d,c},
\end{equation*}
where $m=\ulcorner k/2\urcorner$ and $X_{i,j,d}=0$ if $i\leq 0$ or $i> N$, or $j\leq 0$ or $j> N$.

Each convolutional kernel is corresponding to a linear transformation matrix, and  the output $Y=K*X$ can be reshaped from the multiplication of the transformation matrix with the reshaped $X$.  We use $vec(X)$ to denote the vectorization of $X$. If $X$ is a matrix,  $vec(X)$ is the column vector got  by  stacking the columns of  $X$ on top of one another. If $X$ is a tensor,  $vec(X)$ is the column vector got  by  stacking the columns of the flattening of $X$ along the first index (see \cite{golub2012} for more on flattening of a tensor).   Given a kernel $K$, assume $M$ is the linear transformation matrix corresponding to the kernel $K$, we have
 \begin{equation*}
    vec(Y)=Mvec(X).
\end{equation*}
It is desirable to let $K$ satisfy   that $\|vec(Y)\|\approx \|vec(X)\|$, where $\|\cdot\|$ denotes a certain vector norm,  when training deep convolutional  neural networks \cite{hochreiter2001}.  This is to let the singular values of the linear transformation matrix $M$ corresponding to the kernel $K$ be bounded around $1$.

Given a general kernel $K$ whose size is $k\times k\times g  \times h$ and the input whose size is $ N\times N\times g$, it is not known whether there exists an orthogonal $hN^2\times gN^2$ transformation matrix  corresponding to the $k\times k\times g  \times h$ kernel. It's theoretically difficult to answer this question.  In this paper, we will propose a penalty function for a convolutional kernel to let the singular values of the corresponding transformation matrix be bounded around $1$.
The goal is to minimize the following regularization term
\begin{equation}\label{prob1}
  max\{|\sigma_{max}(M)-1|,|\sigma_{min}(M)-1|\}
\end{equation}
where $M$ is the linear transformation matrix corresponding to kernel $K$.
But the term \eqref{prob1} is hard to minimize directly. For a general matrix $A$, people  let the singular values of $A$ be bounded around $1$ through penalizing the term  $\|A^TA-I\|_2$, where $\|\cdot\|_2$ denotes the spectral norm of a matrix, and $I$ is the identity matrix \cite{kova2008}. In this paper, we will use $\|M^TM-\alpha I\|_F^2$ as the penalty function, where $\|\cdot\|_F$ denotes the Frobenius norm of a matrix,  to let   the singular values of $M$ be bounded around $1$.

As we know, given a matrix, the singular values/eigenvalues are continuous functions depending on the entries of the matrix. We can calculate the partial derivatives of a singular value with respect to the entries, and  let  the singular values of a matrix  be bounded through changing the entries. But the transformation matrix $M$ corresponding to a convolutional kernel is structured, i.e., $M$ has special matrix structure. When changing the entries of $M$, we should preserve the special structure of $M$ such that the updated $M$ can still correspond to the same size convolutional kernel.  In this paper we will show how to preserve the special structure of $M$ when we minimize $\|M^TM-\alpha I\|_F^2$. The modification on $M$ is actually carried out on a special matrix manifold.

There have been papers devoted to enforcing  the orthogonality or spectral norm regularization on the weights of a neural network \cite{brock2017,cisse,miyato2018,yoshida}. The difference between our paper and papers including \cite{brock2017,cisse,miyato2018,yoshida} and the references therein  is about how to handle convolutions. They enforce the constraint directly on the $h\times (gkk)$  matrix reshaped from the kernel $K\in \mathbb{R}^{k\times k\times g  \times h}$,  while we enforce the the  constraint on the transformation matrix $M$ corresponding to the convolution kernel $K$. In  \cite{sedghi2018}, the authors project a convolutional  layer onto the set of layers obeying a bound on the operator norm of the layer and use numerical results to show this is an effective regularizer. A drawback of the method in \cite{sedghi2018} is that projection can prevent the singular values of the transformation matrix being large but can't avoid the singular values to be too small. In \cite{guo2019} a 2-norm regularization method is proposed for convolutional kernels to constrain the singular values of the corresponding transformation matrices. In this paper we propose a Frobenius norm regularization method for convolutional kernels.

The rest of the paper is organized as follows. In subsection \ref{app}, we will introduce the origin of our problem in deep learning applications.
As we have mentioned, the input channels and the output channels maybe more than one so the kernel is usually represented by a tensor $K\in \mathbb{R}^{k\times k\times g  \times h}$.
In Section~\ref{sec:one},  we first consider the case that the numbers of input channels and the output channels are both $1$. We propose the penalty function, calculate the partial derivatives and propose the gradient descent algorithm for this case. In Section \ref{sec:multi}, we propose the penalty function and calculate the partial derivatives for the case of multi-channel convolution.
In Section~\ref{sec:numer},  we present numerical results to show the method is feasible and effective. In Section~\ref{sec:conclu}, we will give some conclusions and point out some interesting work that could be done in future.

\subsection{Applications in deep learning}\label{app}

This problem \eqref{prob1} has important applications in training deep convolutional neural networks.
Convolutional neural network is a very important model of deep learning. A typical convolutional neural network consists of convolutional layers, pooling layers, and fully connected layers. In recent years, deep convolutional neural networks have been applied successfully in many fields, such as face recognition, self-driving cars, natural language understanding and speech recognition. Training the neural networks can be seen as an optimization problem, which is seeking the optimal weights (parameters) by reaching the minimum of loss function on the training data.
This can be described as follows: given a labeled data set $\{(X_i,Y_i)\}_{i=1}^{N}$, where $X_i$ is the input and $Y_i$ is the output, and a given parametric family of functions $\mathbb{F}=\{f(\Theta,X)\}$, where $\Theta$ denotes the parameters contained in the function, the goal of training the neural networks is to find the best parameters $\Theta$ such that $Y_i\approx f(\Theta,X_i)$ for $i=1,\cdots,N$. The practice is to minimize the so called loss function, i.e., $\Sigma_{i=1}^{N}\|Y_i-f(\Theta,X_i)\|$ in certain measure,  on the training data set.

Exploding and vanishing gradients are fundamental obstacles to effective training of deep neural networks \cite{hochreiter2001}. The singular values of the Jacobian of a layer bound the factor by which it changes the norm of the backpropagated signal. If these singular values are all close to $1$, then gradients neither explode nor vanish.

On the other hand, it can help improve the generalizability to let  the singular values of the transformation matrix corresponding to a kernel are bounded around $1$. Although the training of neural networks can be seen as an optimization problem, but the goal of training is not merely to minimize the loss function on training data set. In fact, the performance of the trained model on new data is the ultimate concern. That is to say, after we find the weights or parameters $\Theta$ through minimizing the loss function on training data set, we will use the weights $\Theta$ to get a neural network to predict the output or label for the new input data. Sometimes, the minimum on the training model is reached while the performance on test data is not satisfactory. A concept,  generalizability, is used to describe this phenomenon.
 The generalizability can  be improved through reducing the sensitivity of a loss function against the input data perturbation \cite{goodfellow2013,szegedy2014,tsuzuku2018,zhang,yoshida}.

Therefore, to avoid the exploding/vanishing gradient problem and improve the generalizability of a neural network,  the singular values of the Jacobian of a layer are expected to be close to $1$ in the training process, which can be formulated as a constrained optimization problem. We divide the  weights $\Theta$ of a convolutional neural network into two parts, one is the weights of convolutional layers  and another is the complement.  Assuming the number of convolutional layers is $l$, we use $K_p$ to denote the kernel for the $p$-th convolutional layer and $M_p$ to denote the linear transformation matrix corresponding to $K_p$, and use $W_q, 1\leq q\leq m$ to denote other weight matrices that  belong to other layers. Then researchers in the field of deep learning are interested to consider the following constrained optimization problem:
\begin{equation}\label{opt}
min_{K_1,K_2,\cdots, K_l,W_1,W_2,\cdots,W_m}  \frac{1}{N}\Sigma_{i=1}^{N}\|Y_i-f(K_1,K_2,\cdots, K_l,W_1,W_2,\cdots,W_m, X_i)\|
\end{equation}
\begin{equation*}
\mbox{s.t.} \quad M_p^TM_p\approx I, \quad 1\leq p \leq l, \quad \mbox{and} \quad  W_q^TW_q\approx I,  \quad 1\leq q \leq m,
\end{equation*}
where  $I$ denotes the identity matrix.

\section{penalty function for one-channel convolution}\label{sec:one}

As a warm up, we first focus on the case that  the numbers of input channels and the output channels are both $1$. In this case the weights of the kernel are a $k\times k$ matrix.
Without loss of generality, assuming the data matrix is $N\times N$, we use a $3\times 3$ matrix as a convolution kernel to show the associated transformation matrix. Let $K$ be the convolution kernel,
\begin{eqnarray*}
K=\left(\begin{array}{ccc}
k_{11} & k_{12} & k_{13} \\
k_{21} & k_{22} & k_{23} \\
k_{31} & k_{32} & k_{33}
\end{array}\right).
\end{eqnarray*}
Then the transformation matrix corresponding with the convolution arithmetic is
\begin{eqnarray}\label{conv0}
M=\left(
  \begin{array}{cccccc}
    A_0 & A_{-1} & 0 & 0 & \cdots & 0 \\
    A_1 & A_0 & A_{-1} & \ddots & \ddots & \vdots \\
    0 & A_1 & A_0 & \ddots & \ddots & 0 \\
    0 & \ddots & \ddots & \ddots & A_{-1} & 0 \\
    \vdots & \ddots & \ddots & A_1 & A_0 & A_{-1} \\
    0 & \cdots & 0 & 0 & A_1 & A_0 \\
  \end{array}
\right)
\end{eqnarray}
where
\begin{eqnarray*}
A_0=\left(
  \begin{array}{cccccc}
    k_{22} & k_{23} & 0 & 0 & \cdots & 0 \\
    k_{21} & k_{22} & k_{23} & \ddots & \ddots & \vdots \\
    0 & k_{21} & k_{22} & \ddots & \ddots & 0 \\
    0 & \ddots & \ddots & \ddots & k_{23} & 0 \\
    \vdots & \ddots & \ddots & k_{21} & k_{22} & k_{23} \\
    0 & \cdots & 0 & 0 & k_{21} & k_{22}
  \end{array}
\right),\quad
A_{-1}=\left(
  \begin{array}{cccccc}
    k_{32} & k_{33} & 0 & 0 & \cdots & 0 \\
    k_{31} & k_{32} & k_{33} & \ddots & \ddots & \vdots \\
    0 & k_{31} & k_{32} & \ddots & \ddots & 0 \\
    0 & \ddots & \ddots & \ddots & k_{33} & 0 \\
    \vdots & \ddots & \ddots & k_{31} & k_{32} & k_{33} \\
    0 & \cdots & 0 & 0 & k_{31} & k_{32}
  \end{array}
\right),
\end{eqnarray*}
\begin{eqnarray*}
A_{1}=\left(
  \begin{array}{cccccc}
    k_{12} & k_{13} & 0 & 0 & \cdots & 0 \\
    k_{11} & k_{12} & k_{13} & \ddots & \ddots & \vdots \\
    0 & k_{11} & k_{12} & \ddots & \ddots & 0 \\
    0 & \ddots & \ddots & \ddots & k_{13} & 0 \\
    \vdots & \ddots & \ddots & k_{11} & k_{12} & k_{13} \\
    0 & \cdots & 0 & 0 & k_{11} & k_{12}
  \end{array}
\right).
\end{eqnarray*}
In this case, the transformation matrix $M$ corresponding to the convolutional kernel $K$  is a $N^2\times N^2$ doubly  block banded Toeplitz matrix, i.e., a block banded Toeplitz matrix with its blocks are banded Toeplitz matrices. For the details about Toeplitz matrices, please see references  \cite{chan2007,jin2002}. We will let $n=N^2$ and use $\mathcal{T}$ to denote the set of all matrices like $M$ in \eqref{conv0}, i.e., doubly  block banded Toeplitz matrices with the fixed bandth.

We will use $\|M^TM-\alpha I\|_F^2$ as the penalty function to regularize the convolutional kernel $K$, and calculate $\partial \|M^TM-\alpha I\|_F^2/\partial K_{p,q}$, i.e., the partial derivative of Frobenius norm of $M^TM-\alpha I$ versus  each entry $K_{p,q}$ of the convolution kernel.  Our method provides a new method to calculate the gradient of the penalty function about transformation matrix versus the convolution kernel. People can construct other penalty function about $M$ and get the gradient descent method when training their convolutional networks.
The following lemma is easy but useful in the following derivation.
\begin{lemma}\label{lem1}
The partial derivative of square of Frobenius norm of $A\in\mathbb{R}^{n \times n}$ with respect to each entry $a_{ij}$ is $\partial \|A\|_F^2 /\partial a_{ij}=2a_{ij}$.
\end{lemma}

If an entry $a_{ij}$ of the matrix $A \in\mathbb{R}^{n\times n} $ changes, only the entries belonging to $j$-th row or $j$-th volume of the matrix $A^TA$ are affected.
Actually, we have the following lemma.
\begin{lemma}\label{lem2}
If we use $(A^TA)_{s,t}$ to denote the $(s,t)$ entry of the matrix $A^TA$, then $\partial(A^TA)_{s,t}/ \partial a_{ij}$ is the $(s,t)$ entry of the  matrix
$D=A^T(e_ie_j^T)+(e_je_i^T)A$, where
\begin{eqnarray*}
D=\left( \begin{array}{cccccccc}
   0 & \cdots & \cdots & 0 & a_{i1} & 0 & \cdots & 0 \\
   \vdots & \vdots & \vdots & \vdots & a_{i2} & \vdots & \vdots& \vdots \\
  \vdots & \vdots & \vdots& \vdots & \vdots & \vdots & \vdots & \vdots \\
   0 & \cdots &\cdots & 0 & a_{i,j-1} & 0 & \cdots & 0 \\
   a_{i1} & a_{i2} & \cdots & a_{i,j-1} &2 a_{ij} & a_{i,j+1} & \cdots &  a_{in} \\
   0 & \cdots & \cdots & 0 & a_{i,j+1} & 0 & \cdots &0 \\
   \vdots & \vdots & \vdots & \vdots & \vdots & \vdots & \vdots & \vdots \\
   0 & \cdots & \cdots & 0 & a_{in} & 0 & \cdots & 0
 \end{array}
 \right).
\end{eqnarray*}
\end{lemma}
We have the following formula from lemma~\ref{lem1} and lemma~\ref{lem2}
\begin{eqnarray}\label{derivative1}
 \nonumber  \frac{1}{2}\frac{\partial \|M^TM-\alpha I\|_F^2}{\partial m_{ij}}  &=& \frac{1}{2}\sum_{s,t=1,\cdots,n} \frac{\|M^TM-\alpha I\|_F^2}{(M^TM-\alpha I)_{s,t}}\frac{\partial(M^TM-\alpha I)_{s,t}}{\partial m_{ij}} \\&=& \sum_{t=1,\cdots,n}(M^TM-\alpha I)_{j,t} m_{it}+\sum_{s=1,\cdots,n}(M^TM-\alpha I)_{s,j} m_{is}.
\end{eqnarray}
For a matrix $M\in\mathcal{T}$,  The value of $K_{p,q}$ will appear in different  $(i,j)$ indexes. We use $\Omega$ to denote this index set, i.e., for each $(i,j)\in\Omega$ , we have   $m_{ij}=K_{p,q}$. The chain rule formula about the derivative tells us that, if we want to calculate $\partial \|M^TM-\alpha I\|_F^2/\partial K_{p,q}$, we should calculate  $\partial \|M^TM-\alpha I\|_F^2/\partial m_{ij}$ for all $(i,j)\in\Omega$ and take the sum, i.e.,

\begin{eqnarray}\label{derivative2}
  \nonumber \frac{1}{2} \frac{\partial \|M^TM-\alpha I\|_F^2}{\partial K_{p,q}}&=& \frac{1}{2}\sum_{(i,j)\in\Omega} \frac{\partial \|M^TM-\alpha I\|_F^2}{\partial m_{ij}}\\
  &=& \sum_{(i,j)\in\Omega}(\sum_{t=1,\cdots,n}(M^TM-\alpha I)_{j,t}m_{it}+\sum_{s=1,\cdots,n}(M^TM-\alpha I)_{s,j}m_{is}).
\end{eqnarray}

We summarize the above results as the following theorem. We can use the formula \eqref{derivative3} to carry out the gradient descent method for $\|M^TM-\alpha I\|_F^2$.

\begin{theorem}\label{theo}
Assume  $M\in \mathbb{R}^{n\times n}$ is the doubly  block banded Toeplitz matrix corresponding to the one channel convolution kernel $K\in\mathbb{R}^{k\times k}$. If $\Omega$ is the set of all indexes $(i,j)$ such that $m_{ij}=K_{p,q}$,  we have
\begin{equation}\label{derivative3}
    \frac{1}{2}\frac{\partial \|M^TM-\alpha I\|_F^2}{\partial K_{p,q}}= \sum_{(i,j)\in\Omega}(\sum_{t=1,\cdots,n}(M^TM-\alpha I)_{j,t}m_{it}+\sum_{s=1,\cdots,n}(M^TM-\alpha I)_{s,j}m_{is}).
\end{equation}
\end{theorem}

Theorem~\ref{theo} provides new insight about how to regularize a convolutional kernel $K$ such that singular values of the corresponding transformation matrix are in a bounded interval. We can use the formula \eqref{derivative3} to carry out the gradient type methods.
In future, we can construct other penalty functions to let the transformation matrix corresponding to a convolutional kernel have some prescribed property,  and calculate the gradient of the penalty function with respect to the kernel  as we have done in this paper.

\section{The penalty function and the gradient for multi-channel convolution}\label{sec:multi}
In this section we consider the case of multi-channel convolution. First we show the transformation matrix corresponding to multi-channel convolution.
At each convolutional layer,  we have convolution kernel $K\in \mathbb{R}^{k\times k\times g  \times h}$ and the input $X\in \mathbb{R}^{N\times N\times g}
$; element $X_{i,j,d}$ is the value of the input unit within channel $d$ at row $i$ and column $j$. Each entry of  the output $Y\in \mathbb{R}^{N \times N \times h}$  is produced by
\begin{equation*}
    Y_{r,s,c}= (K*X)_{r,s,c}=\sum_{d\in \{1,\cdots,g\}} \sum_{p\in \{1,\cdots,k\} }\sum_{q\in \{1,\cdots,k\}} X_{r-m+p,s-m+q,d}K_{p,q,d,c},
\end{equation*}
where $X_{i,j,d}=0$ if $i\leq 0$ or $i> N$, or $j\leq 0$ or $j> N$.
By inspection, $vec(Y)=Mvec(X)$, where $M$ is as follows
\begin{eqnarray}\label{conv2}
M=\left(\begin{array}{cccc}
M_{(1)(1)} & M_{(1)(2)} & \cdots& M_{(1)(g)} \\
M_{(2)(1)} & M_{(2)(2)} & \cdots & M_{(2)(g)} \\
\vdots &\vdots &\cdots  &\vdots \\
M_{(h)(1)}& M_{(h)(2)}& \cdots &M_{(h)(g)}
\end{array}\right),
\end{eqnarray}
and each $B_{(c)(d)}\in \mathcal{T}$, i.e., $B_{(c)(d)}$ is a $N^2\times N^2$ doubly  block banded Toeplitz matrix corresponding to the portion $K_{:,:,d,c}$ of $K$ that concerns the effect of the $d$-th input channel on the $c$-th output channel.

Similar as the proof in Section~\ref{sec:one},  we have the following theorem.
\begin{theorem}\label{theo2}
Assume  $M$  is the structured matrix corresponding to the multi-channel convolution kernel $K\in\mathbb{R}^{k\times k\times g \times h}$ as defined in (\ref{conv2}). Given $(p,q,z,y)$, if $\Omega_{p,q,z,y}$ is the set of all indexes $(i,j)$ such that $m_{ij}=k_{p,q,z,y}$,  we have
\begin{equation}\label{derivative4}
    \frac{1}{2}\frac{\partial \|M^TM-\alpha I\|_F^2}{\partial K_{p,q,z,y}}= \sum_{(i,j)\in\Omega}(\sum_{t=1,\cdots,g*N^2}(M^TM-\alpha I)_{j,t}m_{it}+\sum_{s=1,\cdots,g*N^2}(M^TM-\alpha I)_{s,j}m_{is}).
\end{equation}
\end{theorem}
Then the  gradient descent algorithm for the penalty function $\|M^TM-\alpha I\|_F^2$ can be devised, where the number of channels maybe more than one.
We present the detailed gradient descent algorithm for the the penalty function $\|M^TM-\alpha I\|_F^2$ as follows.

\begin{algorithm}\label{alg1}
\noindent \textbf{Gradient Descent for ${\cal R}_\alpha (K) =\|M^TM-\alpha I\|_F^2$.}
\begin{tabbing}
aaaaa \= bbbb\= \kill
1. \> Input: an initial kernel $K\in \mathbb{R}^{k\times k\times g  \times h}$, input size $N\times N\times g$   and learning rate $\lambda$.\\
2. \>While not converged:\\
3. \>\>Compute $G= [\frac{\partial \|M^TM-\alpha I\|_F^2}{\partial k_{p,q,z,y}}]_{p,q,z,y=1}^{k,k,g,h} $, by \eqref{derivative4}; \\
4. \>\> Update $K=K-\lambda G$;\\
5. \>End
\end{tabbing}
\end{algorithm}

\section{Numerical experiments}\label{sec:numer}
The numerical  tests were performed on a laptop (3.0 Ghz and 16G Memory) with MATLAB R2016b.
We use $M$ to denote the transformation matrix corresponding to the convolutional kernel. The largest singular value and smallest singular value of $M$ (denoted as ``$\sigma_{max}(M)$ and $\sigma_{min}(M)$), the iteration steps (denoted as ``iter") are demonstrated to show the effectiveness of our method. The efficiency is related with the step size $\lambda$. According to our experience, the norm of the matrix reshaped from the gradient tensor $G\in \mathbb{R}^{k\times k\times g  \times h}$ in Algorithm \ref{alg1} decreases as the number of iteration steps become larger.  Therefore, we can let the step size $\lambda$ be a small number at first and gradually increase $\lambda$.  In our numerical experiments, for Algorithm \ref{alg1} we  use  the following dynamic adjustment of step size
   \begin{quote}
    if (iter$<10$)\\
        $\lambda=1e-5$;\\
    elseif (iter$<20$)\\
        $\lambda=1e-4$;\\
    else\\
        $\lambda=1e-3$;\\
    end
\end{quote}
Numerical experiments are implemented on extensive test problems. Our method is effective in letting $\sigma_{max}(M)$ and $\sigma_{min}(M)$ be approximate to $1$.
We present the numerical results for some random generated multi-channel convolution kernels.

\begin{figure}[h]
  \centering
  \includegraphics[width=1.00\textwidth]{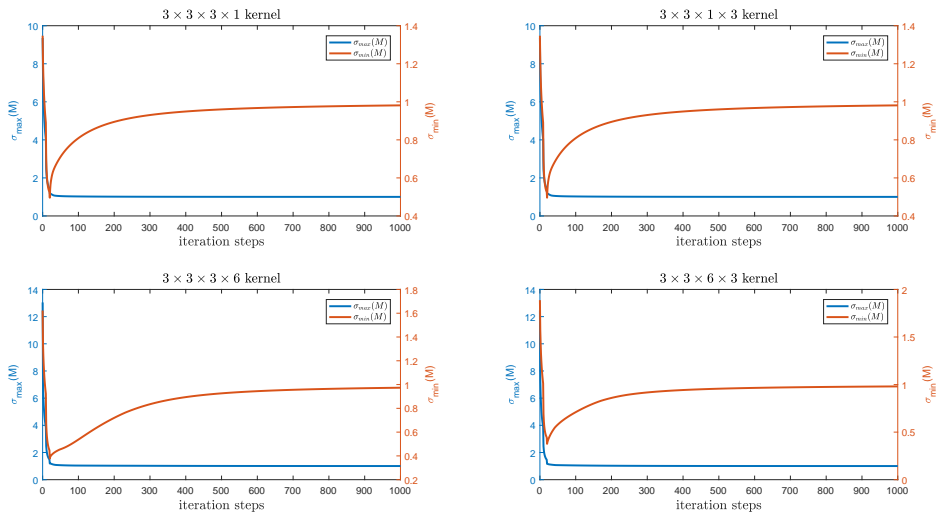}
\caption{\text{\small{Convergence of $\sigma_{max}(M)$ and $\sigma_{min}(M)$ for different kernel sizes}}}
\label{fig1}
  \end{figure}

We start from a random kernel with each entry normally distributed on $[0, 1]$, i.e., in MATLAB, $K$ is generated by the following command
\begin{quote}
    rng(1);\\
    $K= randn(k,k,g,h);$
\end{quote}
We consider kernels of different sizes with $3\times 3$ filters, namely $K\in \mathbb{R}^{3\times 3\times g  \times h}$ for various values of $g,h$.  For each kernel, we use the input data matrix of size   $20\times 20\times g$.  We then minimize $\mathcal{R}_{1} (K) = \|M^TM- I\|_F^2$ using Algorithm \ref{alg1} and we demonstrate the beneficial effect of  decreasing $\sigma_{max}(M)$ while increasing $\sigma_{min}(M)$. We present in Figure~\ref{fig1} the results of $3\times 3\times 3  \times 1$, $3\times 3\times 1  \times 3$, $3\times 3\times 3  \times 6$, and $3\times 3\times 6  \times 3$ kernels. In the figures, we have shown the convergence of $\sigma_{max}(M)$ (blue  line) on the left axis scale and $\sigma_{min}(M)$ (red line) on the right axis scale. From Figure~\ref{fig1}, we see that at the first about 20 steps, $\sigma_{max}(M)$ and $\sigma_{min}(M)$ all decrease and then  $\sigma_{max}(M)$ become very close to 1. Then in the following steps $\sigma_{max}(M)$ become almost unchanged while $\sigma_{min}(M)$ increase from smaller than 1 to be very close to 1.
If the standard of constraining the singular values is not very high, one can stop the gradient descent process after first few steps.

We would like to point out, we have used $\mathcal{R}_{1} (K) = \|M^TM- I\|_F^2$  to do numerical experiments on other random generated examples, including  random kernels with each entry uniformly distributed on $[0, 1]$.
The convergence figures of $\sigma_{max}(M)$ and $\sigma_{min}(M)$ are similar with the subfigures in  Figure~\ref{fig1}.

We noticed that in \cite{guo2019} a 2-norm regularization method about convolutional kernels is proposed. The difference between 2-norm method in \cite{guo2019} and  Frobenius norm method in this paper is needed to investigate further. Now we know that  the computational cost at each iteration step  is the updating of two singular vectors of $M^TM-\alpha I$ for 2-norm method and the computation of the matrix $M^TM-\alpha I$ for Frobenius norm method. The two singular vectors are obtained from two step of power method while the computation of $M^TM-\alpha I$ is done using block matrix multiplication algorithms. We compared the elapsed CPU time of each iteration step for these two  methods and find the time difference  is  little.
The efficiency  of each method, i.e., the needed iteration steps to let $\sigma_{max}(M)$ and $\sigma_{min}(M)$ be bounded in a satisfying interval, is related with the step size $\lambda$. So we can't definitely tell which method is more efficient. But in our extensive numerical experiments,  for each method we test several  different $\lambda$, and choose the ``most" efficient $\lambda$  from these  values of $\lambda$ according to our experience. We use the ``optimal" $\lambda$ for each method to compare the efficiency. Frobenius method needs less iteration steps to let $\sigma_{max}(M)$ and $\sigma_{min}(M)$ be in a designated bounded interval than 2-norm method.

\section{Conclusions}\label{sec:conclu}
In this paper, we provide Frobenius norm method to  regularize the weights of convolutional layers in deep neural networks.
We regularize convolutional kernels to let the singular values of the structured transformation matrix corresponding to a convolutional kernel be close to $1$. We give the penalty function and propose the gradient decent algorithm for the convolutional kernel. We see this method is effective and we will improve it in  future.

In future, we will  continue to  devise other forms of penalty functions to constrain the singular values of structured transformation matrices corresponding to convolutional kernels and apply this type of regularization method into the training of neural networks.
Besides, the details about convergence of the gradient descent method could be focused on. For example, how to choose the optimal parameter $\lambda$ in the algorithm?

\section{Acknowledgements}
The author is grateful to Prof. Qiang Ye for his helpful discussions.


\begin{thebibliography}{99}
\bibitem{brock2017}
Andrew Brock, Theodore Lim, James M Ritchie, and Nick Weston. Neural photo editing with introspective
adversarial networks. In ICLR, 2017.
\bibitem{chan2007}
 R. Chan and X. Jin, An Introduction to Iterative Toeplitz Solvers, SIAM, Philadelphia, 2007.
 \bibitem{cisse}
 Moustapha Cisse, Piotr Bojanowski, Edouard Grave, Yann Dauphin, Nicolas Usunier.
 Parseval Networks: Improving Robustness to Adversarial Examples. In ICML, 2017.
 \bibitem{dumoulin2018}
 Vincent Dumoulin, Francesco Visin. A guide to convolution arithmetic for deep learning. ArXiv, 2018.
 \bibitem{golub2012}
G.-H. Golub and  C.-F. Van Loan, Matrix computations, Johns Hopkins University Press, Baltimore, 2012.
\bibitem{goodfellow2013}
 I. J. Goodfellow, J. Shlens, and C. Szegedy. Explaining and harnessing adversarial examples. In ICLR,
2015.
\bibitem{guo2019}
P. Guo, Q. Ye. On Regularization of Convolutional Kernels in Neural Networks, ArXiv 2019.

\bibitem{hochreiter2001}
S. Hochreiter, Y. Bengio, P. Frasconi, J. Schmidhuber, et al. Gradient flow in recurrent nets: the difficulty of
learning long-term dependencies, In Field Guide to Dynamical Recurrent Networks, IEEE Press, 2001.
\bibitem{jin2002}
 X. Jin, Developments and Applications of Block Toeplitz Iterative Solvers, Science Press, Beijing, 2002.

\bibitem{kova2008}
Kova$\breve{c}$evi$\acute{c}$, Jelena and Chebira, Amina. An introduction to frames, Now Publishers Inc, Boston,
2008.
\bibitem{miyato2018}
Takeru Miyato, Toshiki Kataoka, Masanori Koyama, Yuichi Yoshida. Spectral Normalization for Generative Adversarial Networks. In ICLR, 2018.
\bibitem{sedghi2018}
Hanie Sedghi, Vineet Gupta and Philip M. Long. The Singular Values of Convolutional Layers. In ICLR, 2019.
\bibitem{stewart}
G. W. Stewart. Matrix Algorithms: Volume II. Eigensystems, SIAM, 2001.
\bibitem{szegedy2014}
C. Szegedy, W. Zaremba, I. Sutskever, J. Bruna, D. Erhan, I. J. Goodfellow, and R. Fergus. Intriguing
properties of neural networks. In ICLR, 2014.
\bibitem{tsuzuku2018}
Y. Tsuzuku, I. Sato,  and M. Sugiyama.
Lipschitz-Margin Training: Scalable Certification of Perturbation Invariance for Deep Neural Networks.
In NIPS, 2018.
\bibitem{zhang}
C. Zhang, S. Bengio, M. Hardt, B. Recht, and O. Vinyals. Understanding deep learning requires rethinking generalization. In ICLR, 2017.
\bibitem{yoshida}
Yuichi Yoshida, Takeru Miyato. Spectral Norm Regularization for Improving the Generalizability of Deep Learning, ArXiv 2017.




\end{thebibliography}
\end{document}